\documentclass{article}
\usepackage{jmlr2e}
\usepackage{epsfig,graphicx,subfigure}
\usepackage{amsfonts,amsmath}
\usepackage{latexsym}
\usepackage{enumitem}

\newcommand{\shrink}[1]{}

\newcommand{\reg}{\operatorname{reg}}
\newcommand{\wreg}{\operatorname{wreg}}
\newcommand{\Reg}{\operatorname{Reg}}

\newcommand{\one}[1]{\mathbf{1} \left({#1}\right)}
\newcommand{\I}{\mathbf{1}}

\newcommand{\Learn}{\operatorname{Learner}}
\newcommand{\Costing}{\operatorname{Costing}}
\newcommand{\OT}{\operatorname{Offset\ Tree}}
\newcommand{\BO}{\operatorname{Binary\ Offset}}

\newtheorem{thm}{Theorem}[section]

\newtheorem{defn}{Definition}[section]

\newcommand{\E}{\mathbf{E}}
\renewcommand{\Pr}{\mathbf{Pr}}
\renewcommand{\texttt}{\textrm}

\def\meno{\medskip\noindent}

\usepackage[ruled,vlined]{algorithm2e}

\title{The Offset Tree for Learning with Partial Labels}

\author{\name Alina Beygelzimer \email{beygel@yahoo-inc.com} \\ 
\addr Yahoo Research \thanks{work done while at IBM Research}
\AND
\name John Langford  \email{jcl@microsoft.com} \\
\addr Microsoft Research \thanks{work done while at Yahoo! Research}
} 

\editor{}

\begin{document}

\maketitle

\begin{abstract}
We present an algorithm, called the $\OT$, for learning to make
decisions in situations where the payoff of only one choice is
observed, rather than all choices.  The algorithm reduces this setting
to binary classification, allowing one to reuse of any existing, fully
supervised binary classification algorithm in this partial
information setting.  We show that the Offset Tree is an optimal
reduction to binary classification.  In particular, it has regret at
most $(k-1)$ times the regret of the binary classifier it uses (where
$k$ is the number of choices), and no reduction to binary
classification can do better.  This reduction is also computationally
optimal, both at training and test time, requiring just $O(\log_2 k)$
work to train on an example or make a prediction.

Experiments with the $\OT$ show that it generally performs better than
several alternative approaches.
\end{abstract}

\section{Introduction}\noindent
This paper is about learning to make decisions in partial feedback settings
where the payoff of only one choice is observed rather than all
choices.

As an example, consider an internet site recommending ads or other
content based on such observable quantities as user history and search
engine queries, which are unique or nearly unique for every decision.
After the ad is displayed, a user either clicks on it or not.  This
type of feedback differs critically from the standard supervised
learning setting since we don't observe whether or not the user would have
clicked had a different ad beed displayed instead.

In an online version of the problem, a policy chooses which ads to
display and uses the observed feedback to improve its future ad
choices.  A good solution to this problem must explore different
choices and properly exploit the feedback.  

The problem faced by an internet site, however, is more complex.  They
have observed many interactions historically, and would like to
exploit them in forming an initial policy, which may then be improved
by further online exploration.  Since
exploration decisions have
already been made, online solutions are not applicable.
To properly use the data, we need
\emph{non-interactive} methods for learning with partial feedback.

This paper is about constructing a family of algorithms for
non-interactive learning in such partial feedback settings.  Since any
non-interactive solution can be composed with an exploration
policy to form an algorithm for the online learning setting, the
algorithm proposed here can also be used online.  Indeed, some of our
experiments are done in an online setting.

\subsection*{Problem Definition}
\noindent
Here is a formal description of non-interactive data generation:
\begin{enumerate}
\item Some unknown distribution $D$ generates a feature
  vector $x$ and a vector $\vec{r}=(r_1,r_2,...,r_k)$, where 
  $r_i \in [0,1]$ is the reward of the $i$-th
  action, $i\in\{1,\ldots, k\}$. Only $x$ is revealed to the learner.
\item An existing policy chooses an action $a\in\{1,\ldots,k\}$.
\item The reward $r_a$ is revealed. 
\end{enumerate}
The goal is to learn a policy $\pi: X\rightarrow \{1,\ldots,k\}$ 
for choosing action $a$ given $x$, with the goal of maximizing the expected 
reward with respect to $D$, given by
$$\eta(\pi,D) = \E_{(x,\vec{r})\sim D}\left[r_{\pi(x)}\right].$$
We call this a \emph{partial label problem} (defined by) $D$.

\subsection*{Existing Approaches}
Probably the simplest approach is to regress on the reward $r_a$ given $x$ and
$a$, and then choose according to the largest predicted reward given a new $x$. 
This approach reduces the partial label problem to 
a standard regression problem.

A key technique for analyzing such a reduction is \emph{regret analysis}, which
bounds the ``regret'' of the resulting policy in terms of
the regressor's ``regret'' on the problem of
predicting $r_a$ given $x$ and $a$.  
Here \emph{regret} is
the difference between the largest reward that can be achieved
on the problem
and the reward achieved by the predictor; or---defined in terms of
losses---the difference between the incurred loss and the smallest
achievable loss.  One analyzes excess loss (i.e., regret) 
instead of absolute loss so that the bounds apply to inherently noisy 
problems.  It
turns out that the simple approach above has regret that scales with
the square root of the regressor's regret (see section~\ref{sec:simple} 
for a proof). 
Recalling that the latter is upper bounded by 1, this is
undesirable.

Another natural approach is to use the technique in~\cite{Zadrozny}.
Given a distribution $p(a)$ over the actions given $x$,
the idea is to transform each partial label example 
$\left(x,a,r_a,p(a)\right)$ into an {importance weighted}
multiclass example $\left(x,a,r_a/p(a)\right)$,
where $r_a/p(a)$ is the cost of not predicting label $a$ on input $x$.
These examples are
then fed into any importance weighted multiclass classification algorithm,
with the output classifier used to make future predictions.
Section~\ref{sec:simple} shows that when $p(a)$ is uniform, 
the resulting regret on the original 
partial label problem is bounded 
by $k$ times the importance weighted multiclass regret, 
where $k$ is the number of choices.
The importance weighted multiclass classification problem can, in turn,
be reduced to binary classification, but all known conversions 
yield worse bounds than the approach presented in this paper.

\shrink{
When reduced to binary classification using the
standard all-pairs reduction~\cite{all-pairs}, we get a bound of
$k(k-1)$ times the binary regret.  Other multiclass to binary
reductions can improve the dependence on $k$, but they all yield worse
results than the approach presented in the paper.
}

\subsection*{Results}\noindent 
We propose the $\OT$ algorithm for
reducing the partial label problem to binary classification, allowing one
to reuse any existing, fully supervised 
binary classification algorithm for the partial label problem.

The $\OT$ uses the following trick, which is easiest to understand 
in the case of $k=2$
choices (covered in section~\ref{s:two}).  When the observed reward
$r_a$ of choice $a$ is low, we essentially pretend that the other
choice $a'$ was chosen and a different reward $r_{a'}'$ was observed.
Precisely how this is done and why, is driven by the
regret analysis. 
This basic trick is composable in a binary tree structure for $k>2$,
as described in section~\ref{sec:offset-tree}.  
\shrink{Using an intuition from
bandit algorithms, online regret is minimized by minimizing $r_a/p(a)$.
The symmetrization trick above reduces $r_a$, while 
$p(a)$ is increased by using the conditional probability $p'(a) =
p(a)/(p(a)+p(a'))$.}

The $\OT$ achieves computational efficiency in two ways:
First, it improves the dependence on $k$ from $O(k)$ to 
$O(\log_2 k)$.  It is also an oracle algorithm,
which implies that it can use the implicit optimization in existing
learning algorithms rather than a brute-force enumeration over
policies, as in the Exp4 algorithm~\cite{EXP4}.
We prove that the $\OT$ policy regret is bounded by
$k-1$ times the regret of the binary classifier in solving the
induced binary problems.  Section~\ref{sec:lower-bound} shows 
that no reduction can provide a better guarantee,
giving the first nontrivial lower bound for
learning reductions.  Since the bound is tight
and has a dependence on $k$, 
it shows that the partial label problem
is inherently different from standard fully supervised learning
problems like $k$-class classification.

Section~\ref{sec:simple} analyzes 
several alternative approaches.  An empirical comparison of these
approaches is given in section~\ref{sec:experimental}.

\subsection*{Related Work}\noindent The problem considered here is
a non-interactive version of the contextual bandit
problem (see \cite{Auer,EXP4,EMM,Robbins,Woodruff} for background 
on the bandit problem).  
The interactive version has been analyzed under various additional
assumptions~\cite{Banditron,Kulkarni,Epoch-Greedy,hierarchy_bandit,
Associate,WKP},
including payoffs as a linear function of the side
information~\cite{ABL,Auer}.  The Exp4 algorithm~\cite{EXP4} has a
nice assumption-free analysis.  However, it is intractable when the
number of policies we want to compete with is large.  It also relies on 
careful control of the action choosing distribution, and thus cannot be
applied to historical data, i.e., non-interactively.

Sample complexity results for policy evaluation in reinforcement
learning~\cite{RL} and contextual bandits~\cite{Epoch-Greedy} show
that Empirical Risk Minimization type algorithms can find a good
policy in a non-interactive
setting.  The results here are mostly orthogonal to these 
results, although we do show in section~\ref{sec:sc} that a constant
factor improvement in sample complexity is possible using the offset
trick.

The Banditron algorithm~\cite{Banditron} deals with a similar setting
but does not address several concerns that the $\OT$ addresses: (1)
the Banditron requires an interactive setting; (2) it deals with a
specialization of our setting where the reward for one choice is $1$,
and $0$ for all other choices; (3) its analysis is further specialized
to the case where linear separators with a small hinge loss exist; (4)
it requires exponentially in $k$ more computation; (5) the Banditron
is not an oracle algorithm, so it is unclear, for example, how to
compose it with a decision tree bias.

Transformations from partial label problems to fully supervised
problems can be thought of as learning methods for dealing with sample
selection bias~\cite{bias}, which is heavily studied in Economics and
Statistics.

\section{Basic Definitions}\label{basic}
\noindent
This section reviews several basic learning problems and the 
$\Costing$ method~\cite{costing} used in the construction.

A \emph{$k$-class classification} problem is defined by a distribution
$Q$ over $X\times Y$, where $X$ is an arbitrary feature space and 
$Y$ is a label space with $|Y|=k$.  
The goal is to learn a classifier $c: X\rightarrow Y$ 
minimizing the \emph{error rate} on $Q$, 
$$e(c,Q) = \Pr_{(x,y)\sim Q}[{c(x)\neq y}] = \E_{(x,y)\sim Q}[\,\I({c(x)\neq y})],$$
given training examples of the form $(x,y)\in X\times Y$.
Here $\I(\cdot)$ is the indicator function which evaluates to 
1 when its argument 
is true, and to 0 otherwise.

{Importance weighted classification} is a generalization where
some errors are more costly than others.
Formally, an \emph{importance weighted classification} problem is defined by 
a distribution $P$ over $X\times Y \times [\,0,\infty)$.
Given training examples of the form $(x,y,w)\in X\times Y \times [\,0,\infty)$,
where $w$ is the cost associated with mislabeling $x$,
the goal is to learn a classifier $c: X\rightarrow Y$ 
minimizing the \emph{importance weighted loss} on $P$,
$\E_{(x,y,w)\sim P}[w \cdot \I({c(x)\neq y})]$.

A folk theorem~\cite{costing} says that for any importance weighted
distribution $P$, there exists a constant $\overline{w} =
\E_{(x,y,w)\sim P}[w]$ such that for any classifier $c: X\rightarrow
Y$,
\[ 
\E_{(x,y)\sim Q}[\,\I(c(x)\neq y)] = \frac{1}{\overline{w}}
\E_{(x,y,w)\sim P}[w \cdot \I(c(x)\neq y)],
\]
where $Q$ is the distribution over $X\times Y$ defined by
$$Q(x,y,w) = \frac{w}{\overline{w}}P(x,y,w),$$ marginalized over $w$.
In other words, choosing $c$ to minimize the error rate under $Q$ is
equivalent to choosing $c$ to minimize the importance weighted loss under $P$.

The $\Costing$ method~\cite{costing} can be used to resample the 
training set drawn from $P$ using rejection sampling on 
the importance weights (an
example with weight $w$ is accepted with probability proportional to $w$), 
so that the resampled set is effectively drawn from $Q$.
Then, any binary classification algorithm can be run on 
the resampled set to optimize the importance weighted 
loss on $P$.

$\Costing$ runs a base classification
algorithm on multiple draws of the resampled set, and averages over the learned
classifiers when making importance weighted predictions (see \cite{costing} 
for details). To simplify the analysis, we do not actually have to 
consider separate classifiers.  We can simply augment the feature space
with the index of the resampled set and then learn a single classifier
on the union of all resampled data.
The implication of this observation is that we can view
$\Costing$ as a machine that maps importance weighted examples to unweighted
examples.  We use this method in 
Algorithms~\ref{alg:Binary_Train} and \ref{alg:Filter_Bandit} below.

\shrink{
Thus, we can convert the importance weighted binary examples formed in
step~(\ref{step:binary-offset}) of the algorithm into binary examples
by re-weighting the underlying distribution according to the
importances, and then apply any binary classification algorithm on the
re-weighted training set.  This conversion can be done using the
$\Costing$ algorithm~\cite{costing}, which alters the underlying
distribution using rejection sampling on the importance weights.
}

\section{The Binary Case}\label{s:two}
\noindent
This section deals with the special case of $k=2$ actions.  We state
the algorithm, prove the regret bound (which is later used for the general $k$
case), and state a sample complexity bound.  For simplicity,
we let the two action choices in this section be $1$ and $-1$.

\subsection{The Algorithm}
The $\BO$ algorithm is a reduction from the
2-class partial label problem to binary classification.  The reduction
operates per example, implying that it can be used either online or
offline.  We state it here for the offline case.  The algorithm
reduces the original problem to binary {importance weighted}
classification, which is then reduced to binary classification using the
$\Costing$ method described above.
A base binary classification algorithm $\Learn$ is used as a subroutine.

The key trick appears inside the loop in 
Algorithm~\ref{alg:Binary_Train}, where importance weighted binary
examples are formed.  The offset of $1/2$ changes the range of
importances, effectively reducing the variance of the induced problem.
This trick is driven by the regret analysis in 
section~\ref{binary-regret-theorem}.

\SetAlFnt{\normalsize}
\begin{algorithm}
\dontprintsemicolon
\caption{\label{alg:Binary_Train}
$\BO$ (binary classification algorithm
$\Learn$, 2-class partial label dataset $S$)}
set $S' = \emptyset$ \;
\For{\emph{each $(x,a,r_a,p(a))\in S$}}{
Form an importance weighted example 
\begin{align*}
\quad (x,y,w) = \left(x,\mbox{sign}\left(a\left( r_a - 1/2 \right)\right),\frac{1}{p(a)}\left|r_a - 1/2\right| \right).
\end{align*} 
\;
Add $(x,y,w)$ to $S'$. \;
}
{\bf return} $\Learn(\Costing(S'))$.
\end{algorithm}


\subsection{Regret Analysis}\label{binary-regret-theorem}
This section proves a regret transform theorem for the
$\BO$ reduction. 
Informally, \emph{regret} measures how well a predictor
performs compared to the best possible predictor on the same problem.
A \emph{regret transform} shows
how the regret of a base classifier on the induced (binary classification) 
problem controls the
regret of the resulting policy on the original (partial label) problem.
Thus a regret transform bounds only excess loss 
due to suboptimal prediction.

$\BO$ transforms partial label examples into binary examples.
This process implicitly transforms the
distribution $D$ defining the partial label problem into a
distribution $Q_D$ over binary examples, via a distribution
over importance weighted binary examples.  Note that even though
the latter distribution depends on both $D$ and 
the action-choosing distribution $p$, the induced 
binary distribution $Q_D$ depends only on $D$.
Indeed, the unnormalized probability of label 1 given $x$ and $\vec{r}$, according to 
$Q_D$, is 
\begin{align*}
\E_{a\sim p} &\left[\frac{|r_a-{1/2}|}{p(a)}\cdot
\I\left(a(r_a-\frac{1}{2})>0\right) 
\right] \\
& = \I(r_1>\frac{1}{2})\left|r_1-\frac{1}{2}\right| + 
\I(r_{-1}<\frac{1}{2})\left|r_{-1}-\frac{1}{2}\right|,
\end{align*}
independent of $p$.

The \emph{binary regret} of a classifier $c: X\rightarrow \{-1,1\}$ on
$Q_D$ is given by
\[ 
\reg_e(c,{Q}_D) = e(c,{Q}_D) - \min_{c'}
e(c',{Q}_D), \] 
where the min is over all classifiers $c':
X\rightarrow \{1,-1\}$.  The \emph{importance weighted regret} is definited
similarly with respect to the importance weighted loss.

For the $k=2$ partial label case, the policy that a classifier $c$
induces is simply the classifier.  The regret of policy $c$ is
defined as 
$$\reg_\eta(c,D) = \max_{c'} \eta(c',D) - \eta(c,D),$$
where
$$\eta(c,D) = \E_{(x,\vec{r})\sim D}\left[r_{c(x)}\right],$$
is the value of the policy.

The theorem below states that the policy regret 
is bounded by the binary regret.
We find it
surprising because strictly less information is
available than in binary classification.  Note that the lower bound in
section~\ref{sec:lower-bound} implies that no reduction can do better.
Redoing the proof with the offset set to $0$
rather than $1/2$ also reveals that $2\reg_e(c,Q_D)$
bounds the policy regret, implying that the offset trick gives a factor of 2
improvement in the bound.

Finally, note that the theorem is quantified
over all classifiers, which includes the classifier returned by
$\Learn$ in the last line of the algorithm.

\begin{thm}\emph{($\BO$ Regret)\label{ttwo}}
For all $2$-class partial label problems $D$ and all binary
classifiers $c$, 
\begin{align*}
\reg_\eta(c,D) \leq \reg_e(c,Q_D).
\end{align*}
Furthermore, the bound is tight: There exists a distribution $D$ such that, for 
any value $v \in [0,1]$, there exists a classifier $c$ satisfying $v = \reg_\eta(c,D) = \reg_e(c,Q_D)$.
\end{thm}

\begin{proof}
We first bound the partial label regret
of $c$ in terms of importance weighted regret, and then apply known
results to relate the importance weighted regret to binary regret.

Condition what follows on a fixed $x$, taking the expectation over $x$ at the end.  We can assume without loss of generality that $\E_{\vec{r} \sim
D\mid x} [r_1] \geq \E_{\vec{r} \sim D\mid x} [r_{-1}]$.  If $c(x)=1$, the regrets on both sides are $0$, and the claim holds trivially.  Otherwise, $c$ makes a mistake on $x$.

The importance-weighted error of predicting $-1$ on $x$
is given by
\begin{align*}
\E_{\vec{r} \sim D\mid x}\,& \left[ p(1)\cdot \one{r_1 > 1/2}\cdot\frac{1}{p(1)}\left(r_1-\frac{1}{2}\right) + p(-1)\cdot\one{r_{-1} \leq 1/2}\cdot\frac{1}{p(-1)}\left(\frac{1}{2}-r_{-1}\right) \right] \\
& = \E[\one{r_1 > 1/2} \cdot r_1] - \frac{1}{2}\Pr[{r_1 > 1/2} ] 
- \E[\one{r_{-1} \leq 1/2} \cdot r_{-1}] + \frac{1}{2}\Pr[{r_{-1} \leq 1/2} ],
\end{align*}
where expectations and probabilities are with respect to $\vec{r} \sim D\mid x$.

Similarly, the error of predicting $1$ on $x$ is
\begin{align*}
\frac{1}{2}\Pr[{r_1 \leq 1/2}] - \E[\one{r_1 \leq 1/2} \cdot r_1] + 
\E[\one{r_{-1} > 1/2} \cdot r_{-1}] - \frac{1}{2}\Pr[{r_{-1} > 1/2} ].
\end{align*}

The importance-weighted regret of preferring $-1$ to $1$ on $x$ is given by the difference of the two expressions above.  As the four $1/2$-terms cancel out, the importance-weighted regret simplifies to 
\begin{align*}
 \E[\one{r_1 > 1/2} \cdot r_1 + \one{r_1 \leq 1/2} \cdot r_1] - 
\E[\one{r_{-1} > 1/2} \cdot r_{-1} + \one{r_{-1} \leq 1/2} \cdot r_{-1}] = 
\E[ r_1 - r_{-1}].
\end{align*}


Thus the importance weighted regret of the binary classifier
is the policy regret.  
The folk theorem from section~\ref{basic} (see \cite{costing}) says that
the importance weighted regret is bounded by the binary regret, times the
expected importance.  The latter is 
\[
\E_{\vec{r} \sim D|x} \,\E_{a \sim p(a)}\left[ \frac{1}{p(a)} |r_a - 1/2|\right] = \E_{\vec{r} \sim D|x} \left[\, |r_1 - 1/2| + |r_{-1} - 1/2|\, \right]\leq 1,\]
since both $r_1$ and $r_{-1}$ are bounded by 1.  Taking expectation over $x$ finishes 
the first part of the theorem. 

For the second part, notice that the proof of the first part can be
made an equality by having a reward vector $(0,1)$ for each $x$
always, and letting the classifier predict label $1$ with probability
$(1-v)$ over the draw of $x$.
\end{proof}

\shrink{
\subsection{Sample Complexity Bound}
\label{sec:sc}
\shrink{
A sample complexity bound provides a stronger but narrower guarantee
on the effectiveness of a learning algorithm: essentially it proves
that the learning algorithm competes with a specific set of
classifiers after observing some number of examples, assuming all
examples are drawn IID.
}
This section proves a simple sample complexity bound on the performance of
$\BO$.
For ease of
comparison with existing results, we specialize the problem set
to partial label \emph{binary classification} problems where one label
has reward $1$ and the other label has reward $0$.  Note that this is
not equivalent to assuming realizability:  Conditioned on 
$x$, any distribution over reward vectors $(0,1)$ and $(1,0)$ is allowed.  

Comparing the bound with standard results in binary classification
(see, for example, \cite{Tutorial}), shows that the bounds are identical,
while eliminating the offset trick weakens the performance by a factor 
of roughly 2.

When a sample set is used as a distribution, 
we mean the uniform distribution over the
sample set (i.e., an empirical average).

\begin{thm}\emph{($\BO$ Sample Complexity)\label{ttwo:sample}}
Let the action choosing distribution be uniform.
For all partial label binary classification problems $D$ and
all sets of binary classifiers $C$, after observing a set $S$ of $m$ 
examples drawn independently from $D$, 
with probability at least $1-\delta$, 
\begin{align*}
|\eta(c , D) - \eta(c,S)| \leq \sqrt{\frac{\ln|C| + \ln (2/\delta)}{2 m}}
\end{align*}
holds simultaneously for all classifiers $c \in C$.
Furthermore, if the offset is set to $0$, then 
\begin{align*}
|\eta(c , D) - \eta(c,S)| \leq \sqrt{\frac{\ln|C| + \ln (3/\delta)}{m - 2\sqrt{m\ln(3/\delta)}}}.
\end{align*}
\end{thm}
\begin{proof}
First note that for partial label binary classification problems, the
$\BO$ reduction recovers the correct label.
Since all importance weights are $1$, no examples
are lost in converting from importance weighted classification to
binary classification.  Consequently, the 
Occam's Razor bound on the deviations of error rates implies that, 
with probability $1-\delta$, for all classifiers $c \in C$,
$|e(c,Q_D) - e(c,Q_D)| \leq \sqrt{(\ln|C| + \ln (2/\delta))/{2 m}}$,
where the induced distribution $Q_D$ is 
$D$ with the two reward vectors encoded as binary labels.
Observing that $e(c,Q_D) = \eta(c,D)$ finishes the first half of 
the proof.

For the second half, notice that rejection sampling reduces the number of
examples by a factor of two in expectation; and with probability at least
$1-\delta/3$, this number is at least
$m/2 - \sqrt{m \ln (3/\delta)}$.
Applying the Occam's Razor bound with probability of failure $2 \delta /
3$, gives 
$$|e(c,Q_D) - e(c,Q_D)| \leq \sqrt{\frac{\ln|C| +
    \ln (3/\delta)}{m - 2\sqrt{m \ln (3/\delta)}}}.$$
Taking the union bound over the two failure modes 
proves that the above inequality holds with probability 
$1 - \delta$. Observing the equivalence 
$e(c ,Q_D) = \eta(c, D)$ gives us the final result.
\end{proof}

\noindent
The sample complexity bound provides a stronger (absolute) guarantee, but it
requires samples to be independent and identically distributed.
The regret bound, on the other hand, holds per example and applies always.
}

\section{The Offset Tree Reduction}
\label{sec:offset-tree}
\noindent
In this section we deal with the case of large $k$.

\subsection{The Offset Tree Algorithm}
\noindent
The technique in the previous section can be applied repeatedly using
a tree structure to give an algorithm for general $k$.  Consider a
maximally balanced binary tree on the set of $k$ choices,
conditioned on a given observation $x$.  
Every internal node in the
tree is associated with a classification problem of
predicting which of its
two inputs has the larger expected reward.  
At each node, the same offsetting
technique is used as in the binary case described in
section~\ref{s:two}.

For an internal node $v$, let $\Gamma(T_v)$ denote the set of leaves
in the subtree $T_v$ rooted at $v$. 
Every input to a node is either a leaf or a winning choice from
another internal node closer to the leaves.  


\begin{algorithm}
\dontprintsemicolon
\caption{{$\OT$} (binary classification algorithm $\Learn$,
partial label dataset $S$)}
\label{alg:Filter_Bandit}
Fix a binary tree $T$ over the choices\;
\For{\emph{each internal node $v$ in order from leaves to root}}{
Set $S_v = \emptyset$ \;
\For{\emph{each $(x,a,r_a,p(a)) \in S$ such that $a\in \Gamma(T_v)$ 
and all nodes on the path $v\leadsto a$ predict $a$ on $x$}}{
Let $a'$ be the other choice at $v$ and
$$y=\I(a' \ \mbox{comes from the left subtree of $v$})$$\;
{\bf if} $r_a<1/2$, \\
\qquad add $(x,y, \dfrac{1}{p(a)}(1/2 - r_a))$ to $S_v$\;
{\bf else} add $(x,1-y,\dfrac{1}{p(a)}(r_a - 1/2))$ \;
}
Let $c_v = \Learn(\texttt{Costing}(S_v))$ \;
}
{\bf return} $c=\{ c_v \}$
\end{algorithm}
The training algorithm, $\OT$, is given in
Algorithm~\ref{alg:Filter_Bandit}.  The testing algorithm defining the predictor
is given in Algorithm~\ref{alg:Filter_Policy}.

\begin{algorithm}
\dontprintsemicolon
\caption{\texttt{Offset Test} (classifiers $\{c_v\}$, unlabeled example $x$)}\label{alg:Filter_Policy}
{\bf return} unique action $a$ for which every classifier $c_v$
from $a$ to root prefers $a$.
\end{algorithm}

\subsection{The Offset Tree Regret Theorem}
The theorem below gives an extension of Theorem~\ref{ttwo} for general
$k$.  For the analysis, we use a simple trick which allows us
to consider only a single induced binary problem, and thus a single
binary classifier $c$.  The trick is to add the node index as an
additional feature into each importance weighted binary example
created algorithm~\ref{alg:Filter_Bandit}, and then train based upon the union of all
the training sets.  

As in section~\ref{s:two}, 
the reduction transforms a partial label distribution $D$ into
a distribution $Q_D$ over binary examples. 
To draw from $Q_D$,
we draw $(x,\vec{r})$ from $D$, an action $a$ from the action-choosing 
distribution $p$, and apply algorithm~\ref{alg:Filter_Bandit} to transform
$(x,\vec{r},a,p(a))$ into a set of binary examples (up to one 
for each level in the tree) from which we draw uniformly at random.
Note that $Q_D$ is independent of $p$, as explained in the beginning of 
section~\ref{s:two}.

Denote the policy induced by the
Offset-Test algorithm using classifier $c$ by $\pi_c$.
For the following theorem, the definitions of regret are from
section~\ref{s:two}.

\begin{thm} \emph{($\OT$ Regret)}
\label{otr}
For all $k$-class partial label problems $D$, for all binary
classifiers $c$,
\begin{align*}
\reg_{\eta}(\pi_c,D) & \leq \reg_e(c,Q_D)\cdot  
\E_{(x,\vec{r})\sim D}\hskip -.1in \sum_{v(a,a')\in T}\hskip -.17in\big[\, |r_a - \frac{1}{2}| +|r_{a'} - \frac{1}{2}|\, \big] \\
& \leq (k-1)\reg_e(c,Q_D),
\end{align*}
where $v(a,a')$ ranges over the $(k-1)$ internal nodes in $T$,
and $a$ and $a'$ are its inputs determined by $c$'s predictions.
\end{thm}

\meno
{\bf Note:}\quad
Section~\ref{sec:lower-bound} shows that no reduction can give a better
regret transform theorem. With a little bit of side information, however,
we can do better:  The offset minimizing the regret bound turns out to be
the median value of the reward given
$x$.  Thus,
it is generally best to pair choices which tend to have similar rewards.
Note that the algorithm need not know how well $c$ performs on $Q_D$.

The proof below can be reworked with the offset   
set to $0$, resulting in a regret bound which is a factor of $2$
worse.

\meno\noindent
\begin{proof}
We fix $x$, taking the expectation over the draw of $x$ at the end.
The first step is to show that the partial label regret
is bounded by the sum of the importance weighted regrets over the
binary prediction problems in the tree.  We then apply the costing
analysis~\cite{costing} to bound this sum in terms of the binary
regret.

The proof of the first step is by induction on the nodes in the tree.  We
want to show that the sum of the importance weighted
regrets of the nodes in any subtree bounds the regret of the output
choice for the subtree. The hypothesis trivially holds for one-node
trees.

Consider a node $u$ making an importance weighted decision between
choices $a$ and $a'$.  Without loss of generality, $a$ has the larger expected reward.
As shown in the proof of Theorem~\ref{ttwo}, 
the importance weighted binary regret $\wreg_u$ of the classifier's 
decision at $u$ is either $0$ if it predicts $a$, or $\E_{\vec{r}\sim D|x} [r_{a} - r_{a'}]$
if it predicts $a'$.

Let $T_v$ be the subtree rooted at node $v$, and let $a$ be the choice
output by $T_v$ on $x$.
If the best choice in $\Gamma(T_v)$ comes from the subtree $L$
producing $a$, the policy regret of $T_v$ 
is given by 
\begin{align*}
\Reg({T_v}) &= \max_{y \in \Gamma(L)} \E_{\vec{r}\sim D|x} [r_y] - \E_{\vec{r}\sim D|x} [r_a] \\
&= \Reg({L}) 
\leq \sum_{u \in L}\wreg_u \leq \sum_{u\in {T_v}}\wreg_u.
\end{align*}
If on the other hand the best choice comes from 
the other subtree $R$, we have
\begin{align*}
\Reg({T_v}) &= \max_{y \in \Gamma(R)} \E_{\vec{r}\sim D|x} [r_y] - \E_{\vec{r}\sim D|x} [r_a] \\
&= \Reg({R}) + \E_{\vec{r}\sim D|x} [r_{a'}] - \E_{\vec{r}\sim D|x} [r_a] \\
& \leq \sum_{u\in R} \wreg_{u} + \wreg_v \leq \sum_{u\in {T_v}} \wreg_u,
\end{align*}
proving the induction.

The induction hypothesis applied to $T$ tells us that 
$\Reg(T) \leq \sum_{v\in T}\wreg_{v}$.
According to the Costing theorem discussed in section~\ref{basic},
the importance weighted regret is
bounded by the unweighted regret on the resampled distribution,
times the expected importance.
The expected importance of deciding between
actions $a$ and $a'$ is 
\begin{align*}
\E_{\vec{r} \sim D|x} \left[
p(a)\frac{1}{p(a)} | r_a - 1/2| + p(a')\frac{1}{p(a')}|r_{a'} - 1/2| 
\right]
\leq 1
\end{align*}
since all rewards are between 0 and 1.
Noting that $\Reg(T) = \reg_{\eta}(\pi_c,D\,|\,x)$, we thus have
$$\reg_{\eta}(\pi_c,D\,|\,x) \leq (k-1)\reg_e(c,Q_D\,|\,x),$$
completing the
proof for any $x$. Taking the expectation over $x$ finishes the proof.
\end{proof}
\noindent
The setting above is akin to Boosting~\cite{adaboost}:  At each round $t$, 
a booster creates an input distribution $D_t$ and calls an
oracle learning algorithm to obtain a classifier with some error 
$\epsilon_t$ on $D_t$.  The distribution $D_t$ depends on the classifiers 
returned by the oracle in previous rounds. 
The accuracy of the final classifier is analyzed in terms of $\epsilon_t$'s.
The binary problems induced at internal nodes of an offset tree
depend, similarly, on the classifiers closer to the leaves. 
The performance of the resulting partial label policy is analyzed in terms of 
the oracle's performance on these problems.  (Notice that Theorem~\ref{otr}
makes no assumptions on the error rates on the binary problems; in particular, 
it doesn't require them to be bounded away from $1/2$.)

For the analysis, we use the simple trick from the beginning of this 
subsection to consider only a single binary classifier.
The theorem is quantified over {all} classifiers, and thus it holds 
for the classifier returned by the algorithm.
In practice, one can either call the oracle multiple times 
to learn a separate classifier 
for each node (as we do in our experiments), or use iterative
techniques for dealing with the fact that the classifiers are
dependent on other classifiers closer to the leaves.


\shrink{
The fact that classifiers are conditionally dependent on
other classifiers closer to the leaves causes no problem since
we quantify over {all} classifiers and thus we can regard
the learned classifier as fixed. In practice, one can either use
multiple classifiers (as we do in our experiments) or use iterative
techniques for dealing with the cyclic dependence.
}

\section{A Lower Bound}
\label{sec:lower-bound}
\noindent
This section shows that no method for reducing the partial label setting 
to binary classification can do better.
First we formalize a learning reduction which relies upon a binary 
classification oracle.  The lower bound we prove below
holds for {all} such learning reductions.

\begin{defn}\emph{ (Binary Classification Oracle)  A binary classification 
oracle $O$ is a (stateful) program that supports two kinds of queries:
\begin{enumerate}
\item {\bf Advice}.  An advice query $O(x,y)$ consists of a single
example $(x,y)$, where $x$ is a feature vector and $y \in \{ 1, -1\}$ is a
binary label.  An advice query is equivalent to
presenting the oracle with a training example, and has no return value.
\item {\bf Predict}.  A predict query $O(x)$ is made with a
feature vector $x$.  The return value is a binary label.
\end{enumerate}}
\end{defn}
\noindent
All learning reductions work on a per-example basis,
and that is the representation we work with here.
\shrink{
We typically analyze learning reductions with respect to expectations
over examples. However, all learning reductions work on a per-example basis,
and that is the representation we work with here.
}
\begin{defn}\emph{(Learning Reduction)  A learning reduction is a pair of
algorithms $R$ and $R^{-1}$.
\begin{enumerate}
\item The algorithm $R$ takes a partially labeled example $(x,a,r_a,p(a))$
and a binary classification oracle $O$ as input, and forms a (possibly
dependent) sequence of advice queries.
\item The algorithm $R^{-1}$ takes an unlabeled example $x$ and a
binary classification oracle $O$ as input.  It asks a (possibly
dependent) sequence of predict queries, and
makes a prediction dependent only on the oracle's predictions.
The oracle's predictions may be adversarial (and are assumed so by 
the analysis).
\end{enumerate}}
\end{defn}
\shrink{Since $R$ has access to the partially labeled example and $R^{-1}$
does not, its goal is to advise $R^{-1}$ of which action to choose.
The only means of communication it has is via the Binary
Classification Oracle which adversarially chooses to answer evaluation
queries incorrectly.

The error rate of a binary classifier is defined
for any input $x$ as the fraction of advice queries $O(x_i,y_i)$ for
which evaluation queries are incorrect: $O(x_i) \neq y_i$.  More
generally, when there is a distribution over $x$ we take the expected
value of this fraction.
}

\noindent 
We are now ready to state the lower bound.
\begin{thm} For all reductions $(R, R^{-1})$, there exists a partial label 
problem $D$ and an oracle $O$ such that
$$\reg_{\eta}(R^{-1}(O),D) \geq (k-1)\reg_e(O,R(D)),$$
where $R(D)$ is the binary distribution induced by $R$ on $D$, and 
$R^{-1}(O)$ is the policy resulting from $R^{-1}$ using $O$.
\end{thm}

\begin{proof}
The proof is by construction. We choose $D$ to be uniform over
$k$ examples, with example $i$ having 1 in its $i$-th component of
the reward vector, and zeros elsewhere.  The corresponding feature vector
consists of the binary representation of the index with reward 1.
Let the action-choosing distribution be uniform.

The reduction $R$ produces some simulatable sequence of advice calls
when the observed reward is 0.  The oracle ignores all advice calls
from $R$ and chooses to answer all queries with zero error rate
according to this sequence.

There are two cases: Either $R$ observes $0$ reward (with probability
$(k-1)/k$) or it observes reward $1$ (with probability $1/k$).  In the
first case, the oracle has $0$ error rate (and, hence $0$ regret).
In the second case, it has error rate (and regret) of at most $1$.  
Thus the expected error rate of the oracle on $R(D)$ is at most $1/k$.

The inverse reduction $R^{-1}$ has access to only the unlabeled
example $x$ and the oracle $O$.  Since the oracle's answers are
independent of the draw from $D$, the output action has reward
$0$ with probability $(k-1)/k$ and reward $1$ with probability
$1/k$, implying a regret of $(k-1)/k$ with respect to the best policy.
This is a factor of $k-1$ greater than the regret of the oracle,
proving the lower bound.
\end{proof}

\section{Analysis of Simple Reductions}
\label{sec:simple}
\noindent
This section analyzes two simple approaches for reducing partial
label problems to basic supervised learning problems.  These approaches
have been discussed previously, but the analysis is new.

\subsection{The Regression Approach}
\label{sec:regression}
The most obvious approach is to regress on the value of a choice as in
Algorithm~\ref{alg:Bandit_Regression}, and then use the argmax
classifier as in Algorithm~\ref{alg:Max_Policy}.  
Instead of learning a single regressor, we can learn a 
separate regressor for each choice.


\SetAlFnt{\normalsize}
\begin{algorithm}
\dontprintsemicolon
\caption{\label{alg:Bandit_Regression}
{Partial-Regression} (regression algorithm \texttt{Regress}, partial label dataset $S$)}
Let $S' = \emptyset$ \;
\For{each $(x,a,r_a) \in S$}{
Add $((x,a),r_a)$ to $S'$.\;
}
{\bf return} $f = \texttt{Regress}(S')$.
\end{algorithm}

\begin{algorithm}
\dontprintsemicolon
\caption{\label{alg:Max_Policy}
{Argmax} (regressor $f$, unlabeled example $x$)}
{\bf return} $\arg\max_{a}f(x,a)$
\end{algorithm}
\noindent
The squared error of a regressor $f: X\rightarrow \mathbb{R}$       
on a distribution $P$ over $X\times \mathbb{R}$ is denoted by
$$\ell_r(f,P) = \E_{(x,y)\sim P} (f(x)-y)^2.$$
The corresponding regret is given by
$\reg_r(f,P) = \ell_r(f,P) - \min_{f'} \ell_r(f',P)$.  

\noindent 
The following theorem
relates the regret of the resulting predictor to that of the
learned regressor.

\begin{thm}
For all $k$-class partial label problems $D$ and all
squared-error regressors $f$, 
$$
\reg_{\eta}(\pi_f,D)\leq \sqrt{2k\reg_r(f,P_D)},
$$
where $P_D$ is the regression distribution induced by 
Algorithm~\ref{alg:Bandit_Regression} on $D$, and $\pi_f$
is the argmax policy based on $f$.
Furthermore, there exist $D$ and $h$ such that the bound is tight.
\end{thm}
The theorem has a square root, which is undesirable, because the
theorem is vacuous when the right hand side is greater than 1.

\bigskip
\begin{proof}
Let $\pi_f$ choose some action $a$ with
true value $v_{a}=\E_{(x,\vec{r})\sim D}[r_{a}]$.  Some other action
$a^*$ may have a larger expected reward $v_{a^{*}} > v_{a}$.  The
squared error regret suffered by $f$ on $a$ is $\E_{(x,\vec{r})\sim
  D}[(r_{a}-v_{a})^{2}-(r_{a}-f(x,a))^{2}]=(v_{a}-f(x,a))^{2}$.
Similarly for $a^*$, we have regret
$\left(v_{a^{*}}-f(x,a^{*})\right)^{2}$.  In order for $a$ to be
chosen over $a^{*}$, we must have $f(x,a)\geq f(x,a^{*})$.  Convexity
of the two regrets implies that the minima is reached when
$f(x,a)=f(x,a^{*})=\frac{v_{a}+v_{a^{*}}}{2}$, where the regret for
each of the two choices is
$\left(\frac{v_{a^{*}}-v_{a}}{2}\right)^{2}$.  The regressor need not
suffer any regret on the other $k-2$ arms.  Thus with average regret
$\frac{\left(v_{a^{*}}-v_{a}\right)^{2}}{2k}$ a regret of
$v_{a^{*}}-v_{a}$ can be induced, completing the proof of the first part.
For the second part, note that an adversary can play the
optimal strategy outlined above achieving the bound precisely.
\end{proof}

\subsection{{Importance Weighted Classification}}
\label{sub:The-Importance-Weighted}
Zadrozny~\cite{Zadrozny} noted that the partial label problem could
be reduced to importance weighted multiclass classification.  
After Algorithm~\ref{alg:IW_Bandit} creates importance weighted 
multiclass examples, the weights are stripped
using Costing (the rejection sampling on the weights
discussed in Section~\ref{basic}), and then the resulting 
multiclass distribution is converted
into a binary distribution using, for example, 
the all-pairs reduction~\cite{all-pairs}).
The last step is done to get a comparable analysis.

\begin{algorithm}
\dontprintsemicolon
\caption{\label{alg:IW_Bandit}\texttt{IWC-Train}
(binary classification algorithm \texttt{Learn}, partial label dataset $S$)}
Let $S' = \emptyset$ \;
\For{\emph{each $(x,a,p(a),r_a) \in S$}}{
Add $(x, a, \frac{r_a}{p(a)})$ to $S'$. \;
}
{\bf return} 
$\texttt{All-Pairs-Train}\,(\texttt{Learn}, \texttt{Costing} ( S' ) ) $
\end{algorithm}

\noindent
All-Pairs-Train uses a given binary learning algorithm~\texttt{Learn} to
distinguish each pair of classes in the multiclass distribution created 
by Costing.  
The learned classifier $c$ predicts, given $x$ and                        
a distinct pair of classes $(i,j)$, whether class $i$ is more likely than $j$
given $x$. 
At test time, we make a choice using All-Pairs-Test, which takes
$c$ and an unlabeled example $x$, and returns
the class that wins the most pairwise comparisons on $x$, according to $c$.

\begin{algorithm}
\caption{\label{alg:IW_Policy}\texttt{IWC-Test} (binary classifier $c$,
unlabeled example $x$)}
{\bf return} $\texttt{All-Pairs-Test}(c,x)$.
\end{algorithm}

\noindent
A basic theorem applies to this approach.

\begin{thm}
\label{thm:IW_bandit}
For all $k$-class partial label problems $D$
and all binary classifiers $c$, \emph{
\begin{align*}
\reg_{\eta}(\pi_c,D) 
&\leq \reg_e(c,Q_D)  
(k-1) \E_{(x,\vec{c})\sim D}\sum_{a}(1 - c_a) \\
&\leq \reg_e(c,Q_D) (k-1) k,
\end{align*}
}
where $\pi_c$ is the \emph{IWC-Test} policy based on $c$ and 
$Q_D$ is the binary distribution induced by \emph{IWC-Train} on $D$.
\end{thm}
\begin{proof}
The proof first bounds the policy regret in terms of the 
importance weighted multiclass regret. 
Then, we apply known results for the other
reductions to relate the policy regret to binary classification regret. 

Fix a particular $x$. The policy regret of choosing action $a$ over
the best action $a^{*}$ is 
$\E_{r\sim D|x}[r_{a^{*}}]-\E_{r\sim D|x}[r_{a}]$.
The importance weighted multiclass loss of action $a$
is $$\E_{r\sim D|x}\sum_{a'\neq a}\frac{p(a')r_{a'}}{p(a')}=\E_{r\sim D|x}\sum_{a'\neq a}r_{a'}$$
since the loss is proportional to $\frac{1}{p(a')}r_{a'}$ with
probability $p(a')$. This implies the importance weighted regret of
\[
\E_{r\sim D|x}\sum_{a'\neq a}r_{a'}-\E_{r\sim D|x}\sum_{a'\neq a^{*}}r_{a'}
=\E_{r\sim D|x} [ r_{a^{*}}-r_{a}],\]
which is the same as the policy regret. 

The importance weighted regret is bounded by the unweighted regret, times the
expected importance (see \cite{costing}), 
which in turn is bounded by $k$. Multiclass
regret on $k$ classes is bounded by binary regret times 
$k-1$ using the all-pairs reduction~\cite{all-pairs}, which completes the proof.
\end{proof}

\noindent
Relative to the $\OT$, this theorem has an undesirable extra
factor of $k$ in the regret bound.  While this factor is due to the all-pairs
reduction being a weak regret transform, we are aware of no
alternative approach for reducing multiclass to binary classification that in
composition can yield the same regret transform as the $\OT$.

\section{Experimental Results}
\label{sec:experimental}

\begin{table*}[htbp]
\begin{small}
\centering
\begin{tabular}{|l|r|r||c||c|c||c|c||c|} \hline
& \multicolumn{2}{|c||}{Properties} & & \multicolumn{2}{|c||}{Single regressor} & \multicolumn{2}{|c||}{$k$ regressors} & \\ \hline
{Dataset} & $k$ & $m$ & Weighting & M5P & REPTree & M5P &
REPTree & Offset Tree \\ \hline
ecoli & 8  &  336 & 0.3120 & 0.5663 & 0.3376 & 0.3752 & 0.3811 & 0.2311
\\ \hline
flare & 7  & 1388 & 0.1565 & 0.1570 & 0.1685 &  0.1570 & 0.1592 & 0.1506
\\ \hline
glass & 6 &   214 & 0.5938 & 0.6662 & 0.5846 &  0.5800 & 0.6077 & 0.5000
\\ \hline
letter & 25 & 20000 & 0.3546 & 0.6974 & 0.5491 & 0.4456 & 0.5352 & 0.3790
\\ \hline
lymph & 4 & 148 & 0.2953 &  0.5267 & 0.4622 & 0.3422 & 0.3400 &  0.3114
\\ \hline
optdigits & 10 & 5620 & 0.1682 & 0.5426 & 0.4108 & 0.1948 & 0.2956 &  0.1649
\\ \hline
page-blocks & 5 &  5473 & 0.0407 & 0.0590 & 0.0451 & 0.0571 & 0.0465 &  0.0488
\\ \hline
pendigits & 10 & 10992 & 0.1029 & 0.2492 & 0.1840 & 0.1408 & 0.1774 &  0.0976
\\ \hline
satimage & 6  & 6435 & 0.1703 & 0.2027 & 0.1968 &  0.1787 & 0.1878 & 0.1853
\\ \hline
soybean & 19  & 683 & 0.6533 & 0.8824 & 0.7327 & 0.7688 & 0.7473 &  0.5971
\\ \hline
vehicle & 4   & 846 & 0.3719 & 0.6142 & 0.5665 & 0.3886 & 0.4114 &  0.3743
\\ \hline
vowel & 11   & 990 & 0.6403 & 0.9034 & 0.8919 & 0.7440 & 0.8198 &  0.6501
\\ \hline
yeast & 10  & 1484 & 0.5406 & 0.6626 & 0.5679 & 0.5406 & 0.5697 &  0.4904
\\ \hline
\end{tabular}
\caption{Dataset-specific test error rates (see section~\ref{experiments:reductions}).  Here $k$ is the number of choices and $m$ is the number of examples}
\label{T1}
\end{small}
\end{table*}

\noindent
We conduct two sets of experiments.  The first set compares the Offset
Tree with the two approaches from
section~\ref{sec:simple}.  The second compares with the
Banditron~\cite{Banditron} on the dataset used in that paper.

\subsection{Comparisons with Reductions}
\label{experiments:reductions}
Ideally, this comparison would be with a data source in the partial
label setting.  Unfortunately, data of this sort is rarely available
publicly, so we used a number of publicly available multiclass datasets~\cite{UCI} and
allowed queries for the reward ($1$ or $0$ for correct or wrong) of
only one value per example.

\begin{figure}[t]
\centering
\includegraphics[angle=270,width=.4\textwidth]{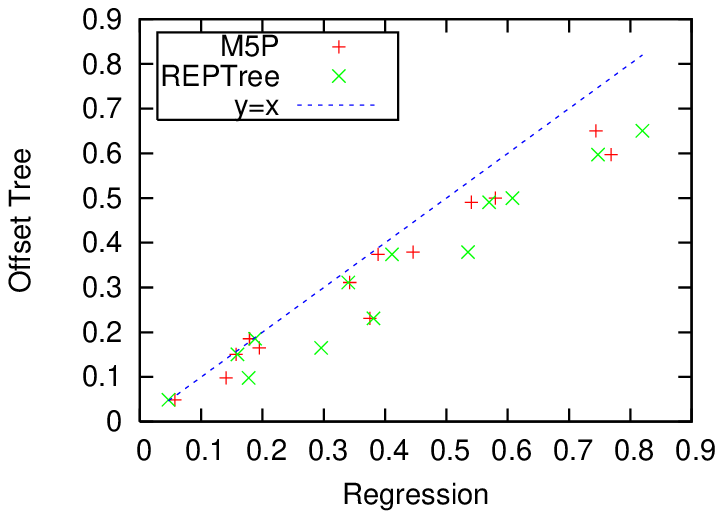}
\includegraphics[angle=270,width=.4\textwidth]{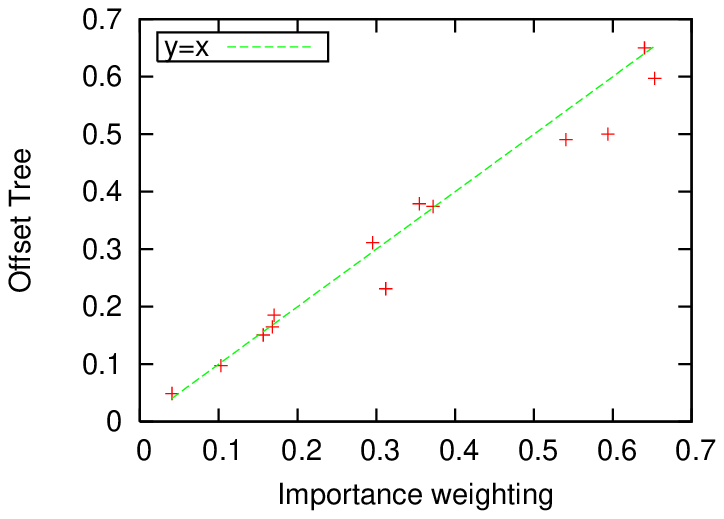}
\caption{\label{fig:comparison} Error rates (in \%)
of $\OT$ versus
the regression approach using two different base regression algorithms
(left) and
$\OT$ versus Importance Sampling (right)
on several different datasets using decision trees
as a base classifier learner.}
\end{figure}

For all datasets, we report the
average result over 10 random splits (fixed for all methods), with
$2/3$ of the dataset used for training and $1/3$ for testing.
Figure~\ref{fig:comparison} shows the error rates (in \%) of
the $\OT$ plotted against the error rates of the regression (left) and
the importance weighting (right).  Decision trees (J48 in
Weka~\cite{weka}) were used as a base binary learning algorithm for
both the $\OT$ and the importance weighting.  For the regression
approach, we learned a separate regressor for each of the $k$ choices.
(A single regressor trained by adding the choice as an additional
feature performed worse.)  M5P and REPTree, both available in
Weka~\cite{weka}, were used as base regression algorithms.

The $\OT$ clearly outperforms regression, in some cases
considerably.  The advantage over importance weighting is
moderate: Often the performance is similar and occasionally it is
substantially better.

We did not perform any parameter tuning because we expect 
that practitioners encountering partial label problems may not have the 
expertise or time for such optimization.  All datasets
tested are included.  Note that although some error rates appear
large, we are choosing among $k$ alternatives and thus an error rate
of less than $1-1/k$ gives an advantage over random guessing.
Dataset-specific test error rates are reported in Table~\ref{T1}.

\subsection{Comparison with the Banditron Algorithm}
The Banditron~\cite{Banditron} is an algorithm for the special case of the 
problem where one of the rewards is $1$ and the rest are $0$.
The sample complexity
guarantees provided for it are particularly good when the correct
choice is separated by a multiclass margin from the other classes.  

We chose the Binary Perceptron as a base classification algorithm since
it is the closest fully supervised learning algorithm to the Banditron.
Exploration was done according to Epoch-Greedy~\cite{Epoch-Greedy} instead
of Epsilon-Greedy (as in the Banditron), motivated by
the observation that the optimal rate of exploration
should decay over time.
The Banditron was tested on one dataset, a 4-class specialization of
the Reuters RCV1 dataset consisting of 673,768 examples.  We use
precisely the same dataset, made available by the authors of \cite{Banditron}.

Since the Banditron analysis suggests the realizable case, and the dataset
tested on is nearly perfectly separable, we also specialized the
$\OT$ for the realizable case.  In particular, in the realizable
case we can freely learn from every observation implying it is unnecessary to
importance weight by $1/p(a)$.  We also specialize Epoch-Greedy to this case by using a
realizable bound, resulting in a probability of exploration that
decays as $1/t^{1/2}$ rather than $1/t^{1/3}$.

The algorithms are compared according to their 
error rate.  For the Banditron, the error rate after one pass
on the dataset was $16.3\%$.  For the realizable $\OT$ method
above, the error rate was $10.72\%$.  For the fully agnostic version of
the $\OT$, the error rate was $18.6\%$.  These results suggest
there is some tradeoff between being optimal when there is arbitrary
noise, and performance when there is no or very little noise.  In the
no-noise situation, the realizable $\OT$ performs substantially
superior to the Banditron.  

\section{Discussion}
\noindent
We have analyzed the 
tractability of learning when only one outcome from a set of $k$ alternatives 
is known, in the reductions setting.  The $\OT$
approach has a worst-case dependence on $k-1$ (Theorem~\ref{otr}), and
no other reduction approach can provide a better guarantee 
(Section~\ref{sec:lower-bound}).
Furthermore, with an $O(\log k)$ computation, the $\OT$ is
qualitatively more efficient than all other known algorithms, 
the best of which are $O(k)$.  
Experimental results suggest that this approach is
empirically promising. 

The algorithms presented here show how to learn from one step of
exploration.  By aggregating information over multiple steps, we can
learn good policies using binary classification methods.  A
straightforward extension of this method to deeper time horizons $T$ is
not compelling as $k-1$ is replaced by $k^T$ in the regret bounds.
Due to the lower bound proved here, it appears that further progress
on the multi-step problem in this framework must come with additional assumptions.

\section{Acknowledgements} 
\noindent
We would like to thank Tong Zhang, Alex Strehl, and Sham Kakade for
helpful discussions.  We would also like to thank Shai Shalev-Shwartz
for providing data and helping setup a clean comparison with the
Banditron.

\appendix
\section{Sample Complexity Bound}
\label{sec:sc}
\shrink{
A sample complexity bound provides a stronger but narrower guarantee
on the effectiveness of a learning algorithm: essentially it proves
that the learning algorithm competes with a specific set of
classifiers after observing some number of examples, assuming all
examples are drawn IID.
}
This section proves a simple sample complexity bound on the performance of
$\BO$.
For ease of
comparison with existing results, we specialize the problem set
to partial label \emph{binary classification} problems where one label
has reward $1$ and the other label has reward $0$.  Note that this is
not equivalent to assuming realizability:  Conditioned on
$x$, any distribution over reward vectors $(0,1)$ and $(1,0)$ is allowed.

Comparing the bound with standard results in binary classification
(see, for example, \cite{Tutorial}), shows that the bounds are identical,
while eliminating the offset trick weakens the performance by a factor
of roughly 2.

When a sample set is used as a distribution,
we mean the uniform distribution over the
sample set (i.e., an empirical average).
\begin{thm}\emph{($\BO$ Sample Complexity)\label{ttwo:sample}}
Let the action choosing distribution be uniform.
For all partial label binary classification problems $D$ and
all sets of binary classifiers $C$, after observing a set $S$ of $m$
examples drawn independently from $D$,
with probability at least $1-\delta$,
\begin{align*}
|\eta(c , D) - \eta(c,S)| \leq \sqrt{\frac{\ln|C| + \ln (2/\delta)}{2 m}}
\end{align*}
holds simultaneously for all classifiers $c \in C$.
Furthermore, if the offset is set to $0$, then
\begin{align*}
|\eta(c , D) - \eta(c,S)| \leq \sqrt{\frac{\ln|C| + \ln (3/\delta)}{m - 2\sqrt{m\ln(3/\delta)}}}.
\end{align*}
\end{thm}
\begin{proof}
First note that for partial label binary classification problems, the
$\BO$ reduction recovers the correct label.
Since all importance weights are $1$, no examples
are lost in converting from importance weighted classification to
binary classification.  Consequently, the
Occam's Razor bound on the deviations of error rates implies that,
with probability $1-\delta$, for all classifiers $c \in C$,
$|e(c,Q_D) - e(c,Q_D)| \leq \sqrt{(\ln|C| + \ln (2/\delta))/{2 m}}$,
where the induced distribution $Q_D$ is
$D$ with the two reward vectors encoded as binary labels.
Observing that $e(c,Q_D) = \eta(c,D)$ finishes the first half of
the proof.

For the second half, notice that rejection sampling reduces the number of
examples by a factor of two in expectation; and with probability at least
$1-\delta/3$, this number is at least
$m/2 - \sqrt{m \ln (3/\delta)}$.
Applying the Occam's Razor bound with probability of failure $2 \delta /
3$, gives
$$|e(c,Q_D) - e(c,Q_D)| \leq \sqrt{\frac{\ln|C| +
    \ln (3/\delta)}{m - 2\sqrt{m \ln (3/\delta)}}}.$$
Taking the union bound over the two failure modes
proves that the above inequality holds with probability
$1 - \delta$. Observing the equivalence
$e(c ,Q_D) = \eta(c, D)$ gives us the final result.
\end{proof}

\noindent
The sample complexity bound provides a stronger (absolute) guarantee, but it
requires samples to be independent and identically distributed.
The regret bound, on the other hand, provides a relative assumption-free guarantee, and thus applies always.


\end{document}